\documentclass[10pt,twocolumn,letterpaper]{article}

\usepackage{cvpr}              
\definecolor{cvprblue}{rgb}{0.21,0.49,0.74}
\usepackage[pagebackref,breaklinks,colorlinks,allcolors=cvprblue]{hyperref}
\usepackage{graphicx}
\usepackage{multirow}
\usepackage{cuted}
\usepackage{capt-of}
\usepackage[table]{xcolor}
\def\confName{CVPR}
\def\confYear{2027}

\title{Bridging Supervision Gaps: A Unified Framework for Remote Sensing Image Change Detection}

\author{
\makebox[\textwidth][c]{
Kaixuan Jiang$^{1}$ \hspace{0.6em}
Chen Wu$^{1}$ \hspace{0.6em}
Zhenghui Zhao$^{1}$ \hspace{0.6em}
Chengxi Han$^{1}$ \hspace{0.6em}
Haonan Guo$^{1}$ \hspace{0.6em}
Hongruixuan Chen$^{2}$
}
\\[0.5em]
$^{1}$Wuhan University \qquad
$^{2}$The University of Tokyo
\\[0.5em]
{\tt\small
\{kaixuan.jiang,chen.wu,zhaozhenghui,chengxihan,haonan.guo\}@whu.edu.cn
}
\\
{\tt\small qschrx@gmail.com}
}

\begin{document}
\maketitle

\begin{strip}
\centering
\includegraphics[scale = 0.45]{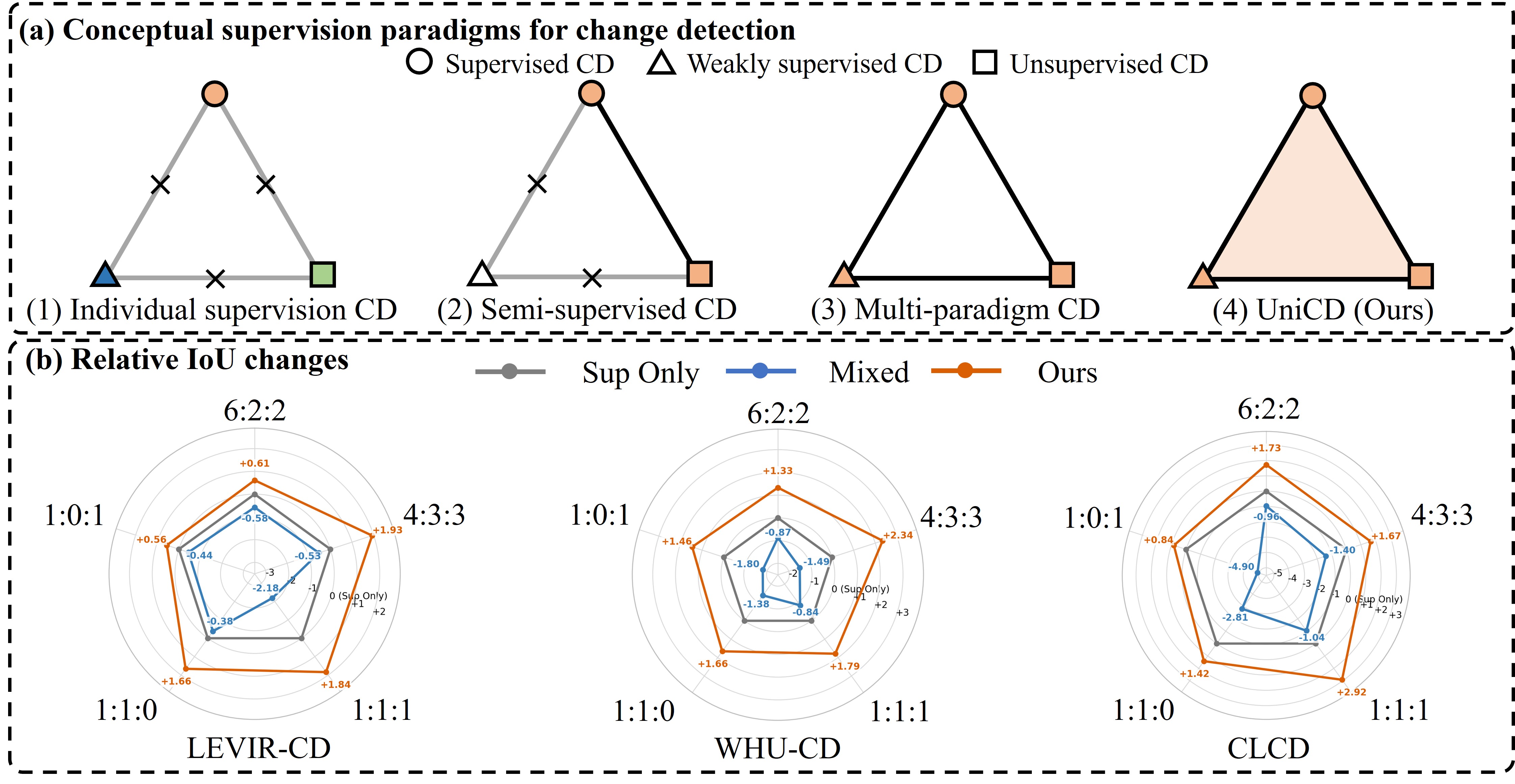}
\captionof{figure}{Comparison of supervision paradigms and mixed-supervision performance. (a) Conceptual comparison of individual supervision methods, semi-supervised CD, FCD-GAN, and the proposed UniCD. (b) Relative IoU changes over the Sup Only baseline under five mixed-supervision ratios on LEVIR-CD, WHU-CD, and CLCD datasets.}
\label{01}
\end{strip}


\begin{abstract}

Change detection (CD) identifies surface changes from bi-temporal remote sensing imagery. Depending on annotation budgets and application contexts, real-world change labels take heterogeneous forms: pixel-level masks, image-level labels, unlabeled samples, or mixtures of these supervision types. However, most existing CD methods target a single supervision paradigm and cannot leverage  complex annotation scenarios. To tackle this challenge, we present UniCD, a unified change detection framework that accommodates supervised, weakly supervised, unsupervised, and mixed-supervision learning within a single network. To bridge supervision granularity gaps, we design Supervision-Anchored Cross-Granularity Prototype Propagation, which constructs class-specific prototypes from pixel-level annotations and transfers reliable change semantics to image-level labeled and unlabeled samples. We further design Multi-source Consistency Rectification to filter unreliable pseudo labels and selectively recover informative pixels through multi-source evidence and cross-view consistency. Experiments on LEVIR-CD, WHU-CD, and CLCD demonstrate that UniCD consistently outperforms conventional mixed training, improving the average IoU by 3.02 percentage points across different mixed-supervision settings, while achieving state-of-the-art performance under individual supervision settings.


\end{abstract}    
\section{Introduction}
\label{sec:intro}

Change detection (CD) identifies land-cover changes from co-registered multi-temporal images ~\cite{wu2024unet} and supports urban monitoring~\cite{lu2004change}, disaster assessment~\cite{lei2026remote}, and agricultural management~\cite{ding2025survey}. With high-revisit constellations, imagery is no longer the bottleneck; labels are. Delineating changes pixel by pixel requires experts to compare two images side by side, so dense change masks remain scarce and expensive. Supervision in practice, however, is rarely all-or-nothing. A realistic archive holds a few image pairs with pixel-level masks, many more with cheap image-level change/no-change tags, and a large pool of unlabeled pairs.

Yet CD methods are still largely organized around individual supervision paradigms~\cite{shi2020change}. We view supervised, weakly supervised, unsupervised, and semi-supervised CD as special cases of a unified problem. The composition of any training set can be represented as a point on a supervision simplex, determined by the fractions of pixel-labeled, image-labeled, and unlabeled pairs, as illustrated in Fig.~\ref{01}a. Conventional single-paradigm settings correspond to the vertices of this simplex, while semi-supervised settings typically lie along its edges. In practice, however, real-world data often contains all three types of supervision and thus occupies the interior of the simplex. Can a single CD model learn from whatever supervision is available, anywhere on the simplex?

The obvious answer is to pool all samples and optimize each with the loss that matches its label. We find that this naive mixing backfires. On LEVIR-CD, WHU-CD, and CLCD under five label mixtures, the pooled model is worse than a model trained on the pixel-labeled subset alone in all 15 settings, by up to 4.9 IoU, as shown in Fig.~\ref{01}b. The extra annotations act as noise rather than information: heterogeneous supervision causes negative transfer.

Related studies in computer vision have explored the joint utilization of heterogeneous annotations in semantic and instance segmentation~\cite{yang2020heterogeneous}, including DSP~\cite{reiss2023decoupled} and WISH~\cite{kweon2025wish}. However, these methods primarily focus on single-image segmentation and do not explicitly model cross-temporal relationships, making them insufficient for directly addressing heterogeneous supervision in CD. Within CD, FCD-GAN~\cite{9fcdgan} supports multiple supervision paradigms through task-specific branches, but does not enable their joint optimization within a unified network. Semi-supervised CD methods~\cite{wang2022reliable} mainly combine pixel-level labeled and unlabeled samples, leaving image-level annotations largely underexplored. Consequently, existing approaches still lack an effective framework for jointly exploiting pixel-level, image-level, and unlabeled data in CD.

We trace this failure to two gaps. The first is a \textbf{granularity gap}: pixel-level masks provide precise spatial semantics, whereas image-level labels only indicate whether a bi-temporal pair has changed, and unlabeled pairs provide no explicit supervision. Consequently, objectives based on class activation maps and consistency may fail to preserve the pixel-level change semantics learned from dense annotations. The second is a \textbf{reliability gap}: weakly supervised and unsupervised learning rely on self-generated pseudo-labels, which are susceptible to pseudo-changes caused by illumination variations, seasonal differences, and image misregistration. Moreover, self-ensembling teachers may reinforce their own prediction errors. These challenges are particularly pronounced in CD, which requires modeling cross-temporal relationships and identifying sparse changed regions. Heterogeneous-supervision methods for single images do not resolve this, because change is a relation between two images and changed pixels are rare.

To address these challenges, we propose UniCD, a unified framework for remote sensing change detection (RSCD) that supports supervised, weakly supervised, unsupervised, and mixed-supervision learning. Specifically, we develop Supervision-Anchored Cross-Granularity Prototype Propagation (SCPP) to bridge the granularity gap by transferring pixel-level semantic knowledge to weakly supervised and unsupervised samples through class-specific prototypes. Furthermore, we introduce Multi-source Consistency Rectification (MCR) to address the reliability gap by screening unreliable pseudo-labels and selectively recovering valid change information. Together, these modules enable more effective utilization of heterogeneous supervision signals within a unified framework. Fig.~\ref{01} illustrates the conceptual differences between existing CD paradigms and the proposed framework. The main contributions are summarized as follows:

\begin{figure*}[!t]
\centering
\includegraphics[scale = 0.53]{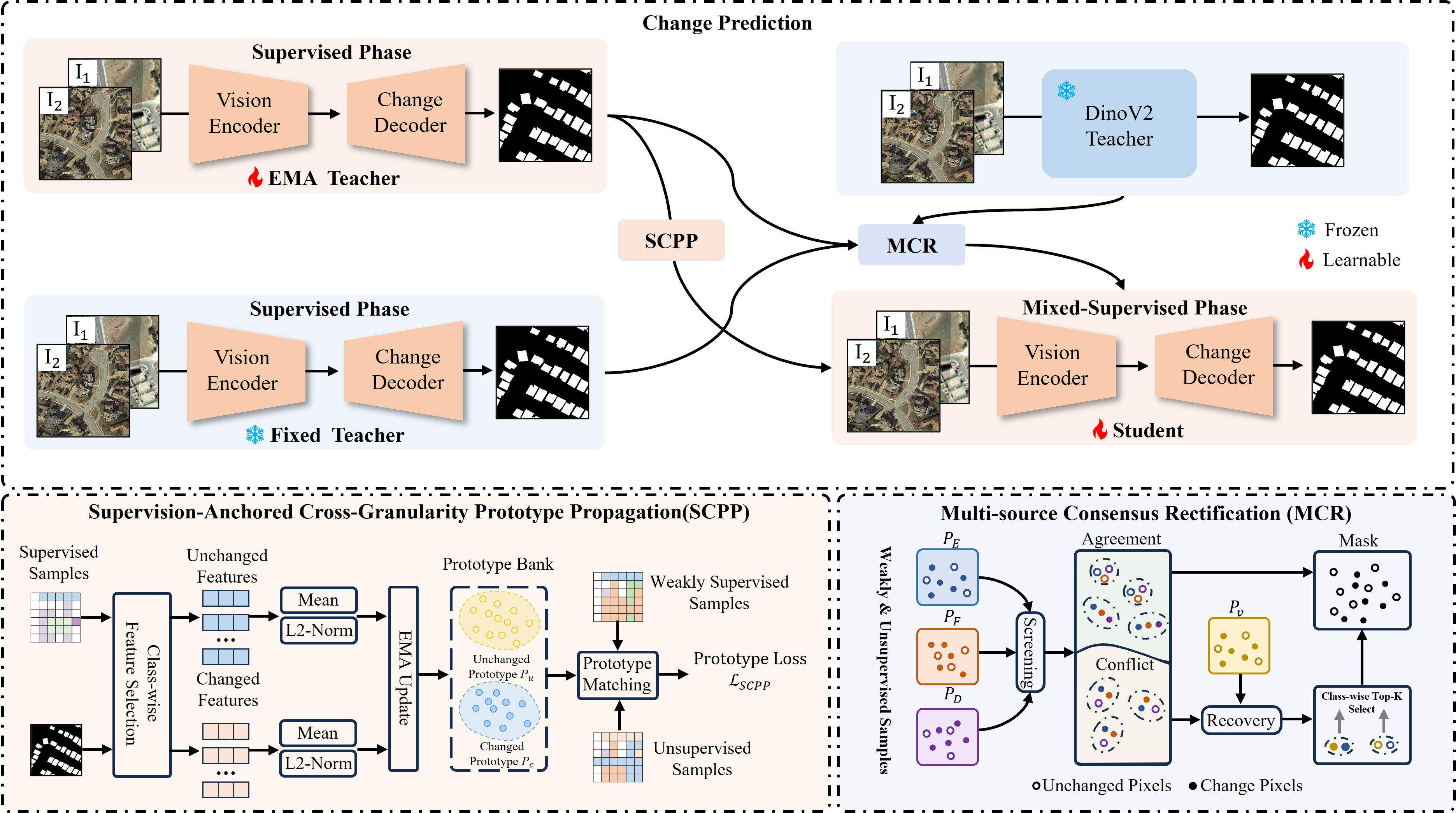}
\caption{Overview of UniCD. The framework supports supervised, weakly supervised, unsupervised, and mixed-supervision learning. SCPP transfers class-specific prototypes from pixel-level annotations to lower-granularity samples, while MCR improves pseudo-label reliability through multi-source consistency screening and selective recovery.}
\label{02}
\end{figure*}

\begin{itemize}

\item We formulate change detection under arbitrary mixtures of pixel-level, image-level, and unlabeled data as learning on a supervision simplex, and show that naive joint training causes negative transfer in all 15 settings we test.

\item  We propose \textbf{UniCD}, to our knowledge the first CD framework that learns jointly from pixel-level, image-level, and unlabeled data in a single network. 

\item It is built on supervision-anchored learning: \textbf{SCPP} anchors the feature space to prototypes estimated only from pixel masks, and \textbf{MCR} anchors the label space through multi-teacher agreement, a foundation-model veto, and budgeted recovery.

\item Extensive experiments on three benchmark datasets demonstrate that UniCD achieves state-of-the-art performance under supervised, weakly supervised, unsupervised, and mixed-supervision settings.

\end{itemize}


\section{Related Work}
\label{sec:formatting}

\subsection{Remote Sensing Change Detection}

Categorized by supervision information, remote sensing CD methods can be divided into three paradigms: supervised, weakly supervised, and unsupervised CD. \textbf{Supervised CD} relies on pixel-level annotations to learn precise change representations, with representative methods including DSAMNet \cite{10dsamnet}, BIT \cite{11bit}, and SAM-CD \cite{13samcd}. Despite their effectiveness, these methods heavily depend on pixel-level annotations, which are costly to obtain and difficult to scale. To reduce annotation requirements, \textbf{weakly supervised CD} utilizes image-level or region-level labels to localize changes through indirect supervision. CS-WSCD employs class activation maps (CAMs) to identify potential change regions \cite{wang2023cs}. Driven by the demand to further reduce reliance on manual annotations, researchers have turned their attention to \textbf{unsupervised CD} that operate entirely without change labels. AnyChange enables zero-shot CD by matching bi-temporal latent features~\cite{anychange}, while S2C learns noise-resistant change representations through contrastive learning~\cite{dings2c}.

Despite steady progress in network architectures and representation learning strategies, most existing remote sensing CD methods are developed under fixed supervision assumptions. This limits their flexibility in practical applications. Such limitations motivate the exploration of more general frameworks that can adaptively handle diverse supervision conditions.

\subsection{Heterogeneous Supervision Learning}

Heterogeneous supervision learning aims to jointly exploit samples annotated at different granularities and has been explored in general computer vision. Omni-DETR \cite{omni2deter} and Omni-RES \cite{omni1} extend omni-supervision to object detection and referring expression segmentation, respectively, by combining fully labeled, weakly labeled, and unlabeled data. Omni-DETR filters teacher predictions through bipartite matching, while Omni-RES uses weak annotations to select and refine teacher pseudo-masks. In semantic segmentation, DSP \cite{reiss2023decoupled} accommodates pixel masks, image-level labels, points, boxes, and unlabeled images through learnable prototype-based segmentation and decoupled contrastive learning. WISH \cite{kweon2025wish} further integrates heterogeneous weak labels, including image tags, points, and boxes, through SAM's prompt latent space and multi-stage matching. These studies demonstrate the potential of exploiting diverse annotations within a unified learning framework.

However, these methods cannot be directly applied to CD, as they focus on single-image segmentation rather than cross-temporal relationships. Existing CD methods provide only partial solutions. Semi-supervised methods, including RCL \cite{wang2022reliable}, C2F-SemiCD \cite{han2024c2f}, and RISL \cite{zuo2024robust}, combine pixel-level labeled and unlabeled data but largely overlook image-level annotations. FCD-GAN \cite{9fcdgan} supports multiple supervision paradigms, including supervised, weakly supervised, regionally supervised, and unsupervised CD, but relies on separate branches and cannot jointly optimize different supervision signals.

In practical remote sensing applications, pixel-level annotated, image-level annotated, and unlabeled samples often coexist within the same dataset. However, jointly exploiting these heterogeneous samples remains challenging due to substantial differences in annotation granularity, spatial constraints, and supervision reliability. Simply combining them during training may introduce semantic inconsistency and optimization interference, limiting the benefits of complementary supervision. Therefore, developing a unified CD framework that accommodates diverse supervision paradigms while effectively coordinating heterogeneous supervision signals remains an important research challenge.
\section{Methodology}

As illustrated in Fig.~\ref{02}, we propose UniCD, a unified framework for CD that supports supervised, weakly supervised, unsupervised, and mixed-supervision learning. Sections~\ref{sec:problem}--\ref{sec:objectives} introduce the unified network and supervision-specific objectives, Sections~\ref{sec:scpp}--\ref{sec:mcr} present SCPP and MCR, respectively, and Section~\ref{sec:optimization} describes the optimization strategy.

\subsection{Problem Formulation and Unified Network}
\label{sec:problem}

We first formulate remote sensing change detection under heterogeneous supervision and then introduce the unified network architecture, including the shared Student network and teacher models.

Given bi-temporal remote sensing images $I_1$ and $I_2$, change detection (CD) aims to predict a pixel-wise binary change map $\hat{Y}=f_{\theta}(I_1,I_2)$. Depending on annotation availability, we consider three disjoint training subsets:
\begin{equation}
\small
\begin{aligned}
\mathcal D &= \mathcal D_s\cup\mathcal D_w\cup\mathcal D_u,\\
\mathcal D_s &= \{(I_1^i,I_2^i,Y^i)\}_{i=1}^{N_s},\\
\mathcal D_w &= \{(I_1^i,I_2^i,y^i)\}_{i=1}^{N_w},\\
\mathcal D_u &= \{(I_1^i,I_2^i)\}_{i=1}^{N_u},
\end{aligned}
\label{eq:unicd_data}
\end{equation}
where $\mathcal D_s$, $\mathcal D_w$, and $\mathcal D_u$ denote the pixel-level, image-level, and unlabeled training subsets, respectively, with $N_s$, $N_w$, and $N_u$ representing their corresponding sample numbers. $Y^i\in\{0,1\}^{H\times W}$ denotes a pixel-level change mask, while $y^i\in\{0,1\}$ represents an image-level label. Their union enables mixed-supervision learning.

\textbf{Unified Student Network.}
UniCD adopts a shared Siamese ResNet-50 encoder and a multi-scale decoder. Specifically, four-level features are extracted as $\{F_t^l\}_{l=1}^{4}=E_{\theta_E}(I_t)$, $t\in\{1,2\}$, and fused through concatenation and projection, i.e., $X^l=\phi_l([F_1^l,F_2^l])$. The fused features are progressively aggregated by three cascaded decoders: $H^1=D_1(X^1,X^2)$, $H^2=D_2(H^1,X^3)$, and $H^3=D_3(H^2,X^4)$. Three independent prediction heads generate spatial logits $Z^k=P_{\theta_k}(H^k)$. The aggregated CAMs and dense predictions are computed as $A=\frac{1}{3}\sum_{k=1}^{3}\operatorname{Softmax}(Z^k)$ and $Z=\frac{1}{3}\sum_{k=1}^{3}Z^k$, respectively. Here, $A$ provides spatial guidance for weakly supervised learning, while $Z$ supports dense prediction. During weakly supervised inference, the CAM prediction is further gated by the image-level prediction from the third head. The final decoder feature $H^3$ is used for subsequent SCPP learning.

\textbf{Teacher Networks.}
UniCD employs an EMA Teacher and a frozen DINOv2 Teacher, with an additional Fixed Teacher introduced for mixed-supervision training. The EMA Teacher shares the Student architecture and is updated after each optimization step using $\theta_{\mathrm{EMA}}\leftarrow\mu\theta_{\mathrm{EMA}}+(1-\mu)\theta_{\mathrm{S}}$, where $\mu$ denotes the momentum coefficient. It is initialized from the Student at the beginning of pure unsupervised training, while in mixed training, both the EMA Teacher and Fixed Teacher are initialized from the supervised warm-up model, with the latter remaining unchanged. The DINOv2 Teacher is an independent frozen pretrained model that extracts complementary semantic features from bi-temporal images. The EMA Teacher provides pseudo labels and cross-granularity guidance for SCPP, while MCR integrates predictions from all three teachers for pseudo-label rectification in mixed-supervision learning.

\subsection{Supervision-Specific Objectives}
\label{sec:objectives}

For pixel-level annotated samples, the unified dense predictions are optimized using pixel-wise cross-entropy, $\mathcal{L}_{\mathrm{sup}}=\operatorname{CE}(Z,Y)$. For image-level annotated samples, weakly supervised learning combines image-level classification, affine feature consistency, and CAM-guided temporal constraints, formulated as $\mathcal{L}_{\mathrm{weak}}=\mathcal{L}_{\mathrm{cls}}+\mathcal{L}_{\mathrm{aff}}+\mathcal{L}_{\mathrm{mask}}$. Specifically, $\mathcal{L}_{\mathrm{cls}}$ applies cross-entropy to the image-level predictions from the original and transformed views, while $\mathcal{L}_{\mathrm{aff}}$ enforces $L_1$ consistency between spatially aligned decoder features. Additionally, $\mathcal{L}_{\mathrm{mask}}$ encourages temporal feature discrepancies in CAM-indicated changed regions while preserving similarity in unchanged regions.

For unlabeled samples, we employ a frozen DINOv2-S/14 model to extract complementary semantic change cues, including cosine feature distances and Jensen--Shannon divergence~\cite{menendez1997jensen} between semantic-center assignments. These cues are combined to generate a change score, from which reliable changed and unchanged pixels are selected using batch-level quantile thresholds. The resulting reliable pixel set $\Omega$ determines where pseudo-label supervision and prediction consistency are applied. Each selected pixel is assigned a DINOv2 reliability weight $q_i^D$, derived from its distance to the corresponding quantile threshold. Meanwhile, an EMA Teacher provides soft probability predictions for consistency regularization. Specifically, the EMA Teacher generates predictions from weakly augmented views to guide the Student on strongly perturbed views through confidence-weighted pseudo-label supervision and prediction consistency:

\begingroup
\small
\setlength{\jot}{1pt}
\begin{equation}
\begin{aligned}
\mathcal{L}_{\mathrm{unsup}}
=&\frac{1}{W_p}
\sum_{i\in\Omega}
q_i^D\alpha_{\hat{Y}_i^D}
\operatorname{CE}(Z_i^S,\hat{Y}_i^D)\\
&+\frac{1}{2W_c}
\sum_{i\in\Omega}
q_i^D
\left\|P_i^S-P_i^E\right\|_2^2,
\end{aligned}
\tag{4}
\end{equation}
\endgroup

where $\Omega$ denotes the reliable pixels selected by DINOv2, and $\hat{Y}_i^D$ represents the corresponding DINOv2-derived hard pseudo label. Here, $\alpha_{\hat{Y}_i^D}$ denotes the class-specific balancing weight, $Z_i^S$ represents the Student logits, and $P_i^S$ and $P_i^E$ denote the Student and EMA Teacher probability vectors, respectively. The DINOv2 reliability weight $q_i^D$ is used to weight both loss terms. The normalization factors are defined as $W_p=\max(1,\sum_{i\in\Omega}q_i^D\alpha_{\hat{Y}_i^D})$ and $W_c=\max(1,\sum_{i\in\Omega}q_i^D)$. The EMA Teacher is progressively updated from the Student parameters.

\subsection{Supervision-Anchored Cross-Granularity Prototype Propagation}
\label{sec:scpp}

Heterogeneous supervision can impose inconsistent semantic constraints: pixel masks specify changed and unchanged regions, whereas image-level labels provide only image-level constraints and unlabeled samples lack explicit spatial annotations. For mixed-supervision training, we introduce Supervision-Anchored Cross-Granularity Prototype Propagation (SCPP) to transfer class-specific knowledge from pixel-annotated samples to the other two subsets.

\textbf{Supervision-anchored prototypes.}
SCPP maintains unchanged and changed prototypes $p_0$ and $p_1$, computed exclusively from supervised masks and the EMA Teacher's final decoder features. For class $c\in\{0,1\}$, we estimate
\begin{equation}
\small
\begin{aligned}
\hat{p}_c
&=
\operatorname{Norm}
\left(
\frac{1}{|\mathcal{I}_c^s|}
\sum_{i\in\mathcal{I}_c^s}
\bar{h}_i^E
\right),\\
\mathcal{I}_c^s
&=\{i:Y_i^r=c\},
\end{aligned}
\tag{5}
\end{equation}
where $Y^r$ denotes the ground-truth mask resized to the decoder feature resolution, $\bar{h}_i^E$ is an $L_2$-normalized EMA Teacher feature, and $\operatorname{Norm}$ denotes $L_2$ normalization. The prototype bank is initialized from $\mathcal{D}_s$ after supervised warm-up and updated only on supervised mini-batches by
\begin{equation}
\small
p_c\leftarrow
\operatorname{Norm}
\left(
\mu_p p_c+(1-\mu_p)\hat{p}_c
\right),
\end{equation}
where $\mu_p$ denotes the momentum coefficient. Thus, weak and pseudo labels never directly update the prototypes.

\textbf{Reliable cross-granularity alignment.}
For a weakly supervised or unlabeled pixel $i$, the EMA Teacher provides a hard pseudo label:
\begin{equation}
\small
\hat{Y}_i^E=\arg\max_c P_{i,c}^E,
\tag{6}
\end{equation}
where $P_i^E$ denotes the EMA Teacher's two-class probability vector. The reliability of these pseudo labels is further assessed by MCR to determine the valid supervision masks. Student features are matched to the detached supervised prototypes using temperature-scaled cosine similarity:
\begin{equation}
\small
s_{i,c}
=
\frac{\operatorname{Norm}(h_i^S)^\top p_c}{\tau_p},
\end{equation}
where $\tau_p$ is the temperature parameter and $h_i^S$ denotes the Student feature. SCPP jointly constrains prototype-space classification and dense change prediction:
\begingroup
\small
\setlength{\jot}{1pt}
\begin{equation}
\begin{aligned}
\mathcal{L}_{\mathrm{SCPP}}
=&\frac{1}{N_m}
\sum_i m_i\operatorname{CE}(s_i,\tilde{Y}_i)\\
&+\frac{1}{N_m}
\sum_i m_i\operatorname{CE}(Z_i^S,\tilde{Y}_i),
\end{aligned}
\tag{7}
\end{equation}
\endgroup
where $s_i=(s_{i,0},s_{i,1})$ denotes the prototype similarity logits, $Z_i^S$ denotes the Student pixel-level logits, and $\tilde{Y}_i=\hat{Y}_i^E$ is the EMA Teacher's hard target. The mask $m_i$ denotes the final valid mask, given by $b_i^w$ for weakly supervised samples and $m_i^{\mathrm{MCR}}$ for unlabeled samples, as defined in Sec.~\ref{sec:mcr}. $N_m=\max(1,\sum_i m_i)$ denotes the number of valid pixels, with a minimum value of one to avoid division by zero.

\begin{figure*}[!t]
\centering
\includegraphics[scale = 0.48]{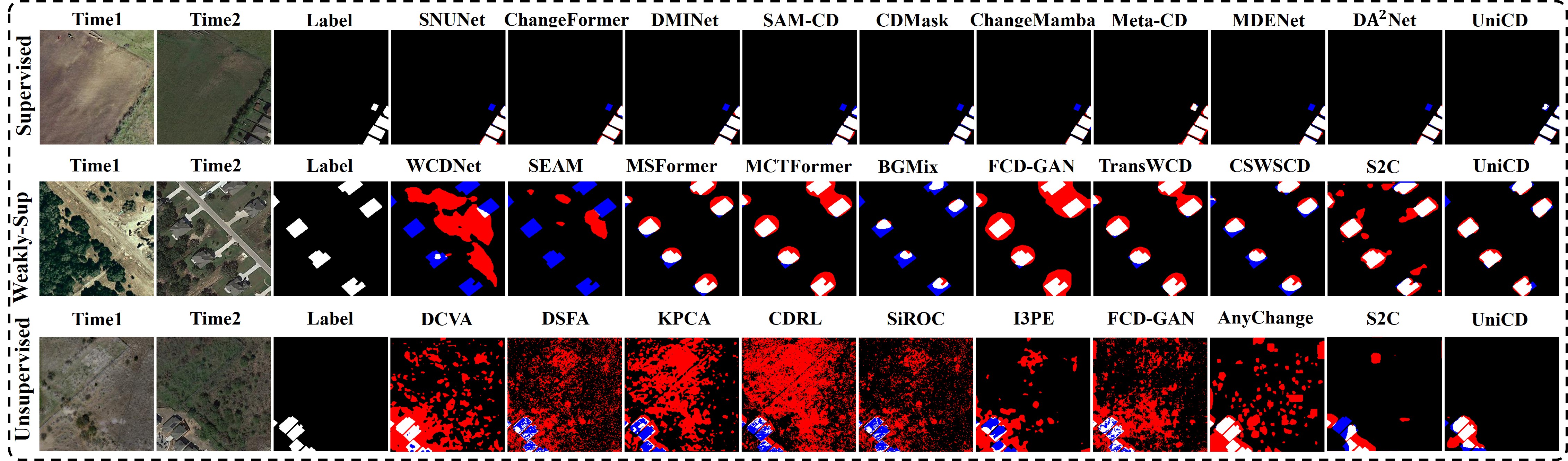}
\caption{\textbf{Qualitative comparison} on LEVIR-CD under supervised, weakly supervised, and unsupervised settings. Red and blue indicate false-positive and false-negative pixels, respectively. UniCD achieves more accurate change localization with fewer false positives and false negatives across all three supervision settings.}
\label{i4}
\end{figure*}

\subsection{Multi-source Consistency Rectification}
\label{sec:mcr}

Although SCPP promotes semantic alignment across heterogeneous supervision, noisy pseudo labels may still degrade training. For mixed-supervision training, we further propose Multi-source Consistency Rectification (MCR), which combines multi-source reliability screening with class-balanced selective recovery to improve pseudo-label quality and supervision coverage.

\textbf{Reliability screening.}
Let $E$ and $F$ denote the EMA Teacher and the Fixed Teacher initialized from the supervised warm-up model, respectively. For each teacher $a\in\{E,F\}$, we define
\begin{equation}
\small
\hat{Y}_i^a=\arg\max_c P_{i,c}^a,
\qquad
q_i^a=\max_c P_{i,c}^a,
\end{equation}
as its predicted class and confidence. For DINOv2, $\hat{Y}_i^D$ denotes its pseudo label, $q_i^D$ denotes the same reliability weight introduced in Sec.~3.2, and $m_i^D$ indicates the validity of its pseudo label. A pixel is retained when both teachers provide confident, consistent predictions. For unlabeled samples, conflicting high-confidence DINOv2 evidence further excludes unreliable pixels:
\begin{equation}
\small
\begin{aligned}
b_i^w
&=
\mathbb{I}
\left[
\min(q_i^E,q_i^F)\geq\tau_T
\land
\hat{Y}_i^E=\hat{Y}_i^F
\right],\\
v_i^D
&=
m_i^D
\mathbb{I}
\left[
q_i^D\geq\tau_D
\land
\hat{Y}_i^D\neq\hat{Y}_i^E
\right],\\
b_i^u
&=
b_i^w(1-v_i^D).
\end{aligned}
\tag{8}
\end{equation}
Here, $\mathbb{I}[\cdot]$ denotes the indicator function, returning 1 when the condition is satisfied and 0 otherwise. The teacher confidence threshold and DINOv2 reliability threshold are set to $\tau_T=0.9$ and $\tau_D=0.75$, respectively. $b_i^w$ and $b_i^u$ denote the valid masks for weakly supervised and unlabeled samples, respectively. The DINOv2 conflict check is applied only to unlabeled samples.

\textbf{Selective recovery.}
Conservative screening may discard reliable pixels when the Fixed Teacher is uncertain. To recover such supervision, we exploit EMA prediction consistency between the original and horizontally flipped views. After inverse alignment, the flipped-view prediction is denoted by $\hat{Y}_i^V$ with confidence $q_i^V$. A previously rejected unlabeled pixel becomes a recovery candidate if both EMA views agree with high confidence, the Fixed Teacher remains uncertain, and no reliable DINOv2 conflict exists:
\begin{equation}
\small
r_i=
(1-v_i^D)
\mathbb{I}
\left[
\min(q_i^E,q_i^V)\geq\tau_T
\land
\hat{Y}_i^E=\hat{Y}_i^V
\land
q_i^F<\tau_T
\right].
\tag{9}
\end{equation}
Candidates are ranked by their minimum cross-view confidence, and only the top $K$ pixels per class and image are recovered, forming $\mathcal{R}_0$ and $\mathcal{R}_1$. For unlabeled samples, the final valid mask is
\begin{equation}
\small
m_i^{\mathrm{MCR}}
=
b_i^u
\lor
\mathbb{I}[i\in\mathcal{R}_0\cup\mathcal{R}_1].
\tag{10}
\end{equation}
For weakly supervised samples, the final valid mask remains $b_i^w$. The recovery budget is limited to approximately 0.5\% of feature-grid pixels per class and image. Previously accepted pseudo labels remain unchanged, while recovered pixels adopt EMA hard predictions. MCR refines the supervision masks used by the existing unlabeled and SCPP losses without introducing an additional loss term.

\subsection{Heterogeneous-Supervision Optimization}
\label{sec:optimization}
To facilitate the joint optimization of heterogeneous supervision, we adopt a two-stage training strategy. It combines supervised warm-up with subsequent joint training to progressively exploit pixel-level, image-level, and unlabeled data. Specifically, during the first 100 epochs, the shared network is trained exclusively on pixel-level annotated samples from $\mathcal{D}_s$ to establish reliable initial representations. After supervised warm-up, the EMA Teacher and Fixed Teacher are initialized from the warmed-up Student, and the supervised prototype bank is initialized using EMA Teacher features extracted from $\mathcal{D}_s$. The Fixed Teacher is retained as a frozen reference for MCR. During the remaining 100 epochs, mini-batches from $\mathcal{D}_s$, $\mathcal{D}_w$, and $\mathcal{D}_u$ are randomly interleaved to jointly optimize the shared Student network. The EMA Teacher is updated after each optimization step, whereas the Fixed Teacher and DINOv2 Teacher remain frozen.

The overall training objective is formulated as
\begin{equation}
\mathcal{L}_{\mathrm{hetero}}=
\begin{cases}
\mathcal{L}_{\mathrm{sup}},
& \mathcal{B}\sim\mathcal{D}_s,\\
\mathcal{L}_{\mathrm{weak}}+\lambda_p\mathcal{L}_{\mathrm{SCPP}}^{w},
& \mathcal{B}\sim\mathcal{D}_w,\\
\mathcal{L}_{\mathrm{unsup}}+\lambda_p\mathcal{L}_{\mathrm{SCPP}}^{u},
& \mathcal{B}\sim\mathcal{D}_u,
\end{cases}
\tag{11}
\end{equation}
where $\mathcal{B}$ denotes a mini-batch sampled from the corresponding training subset. $\mathcal{L}_{\mathrm{weak}}$ and $\mathcal{L}_{\mathrm{unsup}}$ denote the weakly supervised and unsupervised objectives, respectively, while $\mathcal{L}_{\mathrm{SCPP}}^{w}$ and $\mathcal{L}_{\mathrm{SCPP}}^{u}$ represent the corresponding SCPP losses. $\lambda_p$ controls the contribution of SCPP and is set to 0.2. MCR refines pseudo labels and validity masks without introducing an additional loss term.

\section{Experiment}

\subsection{Experimental Setting}

{\setlength{\tabcolsep}{1.5mm}
\begin{table*}[!t]
\footnotesize
\centering
\renewcommand{\arraystretch}{0.7}
\setlength{\tabcolsep}{6pt}
\caption{\textbf{Performance comparison under different mixed-supervision} ratios on LEVIR-CD, WHU-CD, and CLCD datasets, where the ratios denote $\mathcal{D}_{s}:\mathcal{D}_{w}:\mathcal{D}_{u}$. All results are reported in percentages (\%), and the best values are highlighted in bold.}
\label{mixed_ratio}
\begin{tabular}{l|l|cc|cc|cc|cc|cc}
\toprule[1.5pt]
\multirow{2}{*}{\textbf{Dataset}}
& \multirow{2}{*}{\textbf{Method}}
& \multicolumn{2}{c|}{\textbf{6:2:2}}
& \multicolumn{2}{c|}{\textbf{4:3:3}}
& \multicolumn{2}{c}{\textbf{1:1:1}}
& \multicolumn{2}{c}{\textbf{1:1:0}}
& \multicolumn{2}{c}{\textbf{1:0:1}}\\
\cline{3-12}
&
& \textbf{F1} & \textbf{IoU}
& \textbf{F1} & \textbf{IoU}
& \textbf{F1} & \textbf{IoU}
& \textbf{F1} & \textbf{IoU} 
& \textbf{F1} & \textbf{IoU} \\
\midrule[1pt]

\multirow{5}{*}{LEVIR-CD}
& Sup Only
& 91.32 & 84.03
& 90.31 & 82.32
& 88.36 & 79.14 
& 88.36 & 79.14 
& 88.36 & 79.14  \\
& Mixed
& 90.98 & 83.45
& 91.13 & 81.79
& 86.98 & 76.96 
& 88.12 & 78.76
& 88.08 & 78.70 \\

& + SCPP
& 91.53 & 84.38
& 90.99 & 83.47
& 88.78 & 79.83 
& 89.18 & 80.47
& 88.67 & 79.65 \\
& + MCR
& 91.60 & 84.49
& 91.15 & 83.74
& 89.22 & 80.54 
& 89.07 & 80.30
& 88.55 & 79.45 \\

& + SCPP + MCR
& \textbf{91.68} & \textbf{84.64}
& \textbf{91.45} & \textbf{84.25}
& \textbf{89.49} & \textbf{80.98} 
& \textbf{89.38} & \textbf{80.80} 
& \textbf{88.70} & \textbf{79.70}  \\
\midrule

\multirow{5}{*}{WHU-CD}
& Sup Only
& 92.74 & 86.47
& 92.14 & 85.42
& 85.94 & 75.35 
& 85.94 & 75.35
& 85.94 & 75.35\\
& Mixed
& 92.24 & 85.60
& 91.26 & 83.93
& 85.40 & 74.51 
&85.04  & 73.97
&84.76  & 73.55\\
& + SCPP
& 93.20 & 87.27
& 93.31 & 87.46
& 86.78 & 76.65 
& 86.95 & 76.91
&86.16  &75.69 \\
& + MCR
& 93.39 & 87.59
& 93.33 & 87.50
& 86.38 & 76.03 
& 86.14  & 75.66
& 86.40 &76.05\\

& + SCPP + MCR
& \textbf{93.50} & \textbf{87.80}
& \textbf{93.48} & \textbf{87.76}
& \textbf{87.10} & \textbf{77.14}
& \textbf{87.01} & \textbf{77.01}
& \textbf{86.88} & \textbf{76.81} \\
\midrule

\multirow{5}{*}{CLCD}
& Sup Only
& 73.12 & 57.63
& 72.49 & 56.85
& 60.00 & 42.85 
& 60.00 & 42.85 
& 60.00 & 42.85 \\
& Mixed
& 72.35 & 56.67
& 71.34 & 55.45
& 58.97 & 41.81 
& 57.18 & 40.04
& 55.02 & 37.95 \\
& + SCPP
& 74.16 & 58.93
& 73.51 & 58.12
& 61.75 & 44.67
& 60.66 & 43.53
& 60.55 & 43.42 \\

& + MCR
& 74.45 & 59.29
& 73.37 & 57.94
& 60.44 & 43.30 
& 60.50 & 43.37
& 60.41 & 43.27\\

& + SCPP + MCR
& \textbf{74.50} & \textbf{59.36}
& \textbf{73.83} & \textbf{58.52}
& \textbf{62.80} & \textbf{45.77} 
& \textbf{61.37} & \textbf{44.27}
& \textbf{60.81 } & \textbf{43.69} \\
\bottomrule[1.5pt]
\end{tabular}
\end{table*}
}

\begin{table}[!t]
\scriptsize
\centering
\renewcommand{\arraystretch}{1.1}
\setlength{\tabcolsep}{1.1pt}
\caption{\textbf{Fully supervised results} (\%) on LEVIR-CD, WHU-CD, and CLCD. The best values are highlighted in bold.}
\label{supervised_comparison}
\begin{tabular}{l|ccc|ccc|ccc}
\toprule[1.5pt]
\multirow{2}{*}{\textbf{Model}}
& \multicolumn{3}{c|}{\textbf{LEVIR-CD}}
& \multicolumn{3}{c|}{\textbf{WHU-CD}}
& \multicolumn{3}{c}{\textbf{CLCD}} \\
\cline{2-10}
& \textbf{F1} & \textbf{IoU} & \textbf{OA}
& \textbf{F1} & \textbf{IoU} & \textbf{OA}
& \textbf{F1} & \textbf{IoU} & \textbf{OA} \\
\midrule[1pt]
SNUNet$_{\scriptscriptstyle\mathrm{GRSL}}$
& 89.96 & 81.75 & 98.97 & 79.69 & 66.23 & 96.82 & 64.01 & 47.06 & 94.62 \\

ChangeFormer$_{\scriptscriptstyle\mathrm{IGARSS}}$
& 89.10 & 80.34 & 98.89 & 83.77 & 72.07 & 97.84 & 64.16 & 47.24 & 94.84 \\

DMINet$_{\scriptscriptstyle\mathrm{TGRS}}$
& 88.78 & 79.82 & 98.85 & 82.11 & 69.65 & 98.39 & 63.08 & 46.07 & 94.36 \\

SAM-CD$_{\scriptscriptstyle\mathrm{TGRS}}$
& 89.53 & 81.06 & 98.48 & 91.40 & 84.51 & 99.41 & 63.74 & 46.78 & 94.73 \\

CDMask$_{\scriptscriptstyle\mathrm{TGRS}}$
& 90.48 & 82.61 & 99.21 & 91.44 & 84.23 & 99.39 & 67.13 & 50.53 & 94.94 \\

ChangeMamba$_{\scriptscriptstyle\mathrm{TGRS}}$
& 90.37 & 82.43 & 99.03 & 91.96 & 85.11 & 99.45 & 71.71 & 55.89 & 95.98 \\

Meta-CD$_{\scriptscriptstyle\mathrm{TGRS}}$
& 91.20 & 83.82 & \textbf{99.26} & 92.16 & 85.46 & 99.44 & 70.24 & 54.13 & 94.95 \\

MDENet$_{\scriptscriptstyle\mathrm{JSTARS}}$
& 91.33 & 84.05 & 99.11 & 91.92 & 85.05 & 99.35 & 71.53 & 55.68 & 95.70 \\

DA$^{2}$Net$_{\scriptscriptstyle\mathrm{TGRS}}$
& 91.40 & 84.17 & 99.12 & 92.48 & 86.02 & 99.42 & 71.99 & 56.24 & 96.20 \\

\midrule[1pt]
\textbf{UniCD}
& \textbf{91.90} & \textbf{85.02} & 99.17
& \textbf{93.97} & \textbf{88.64} & \textbf{99.52}
& \textbf{76.08} & \textbf{61.38} & \textbf{96.54} \\
\bottomrule[1.5pt]
\end{tabular}
\end{table}

\textbf{Dataset descriptions.} To validate UniCD's performance across various CD tasks, we conduct extensive experiments on three benchmark datasets: LEVIR-CD \cite{18levir}, WHU-CD \cite{19whucd}, and CLCD \cite{20clcd}. LEVIR-CD and WHU-CD primarily focus on building CD in urban scenes, whereas CLCD targets cropland detection in agricultural areas.

\textbf{Implementation details.} All experiments in this paper are implemented on the PyTorch framework \cite{21pytorch} and trained on a single NVIDIA GeForce RTX 3090 GPU. We adopt AdamW \cite{loshchilov2017decoupled} as the optimizer with an initial learning rate of $1 \times 10^{-4}$. For mixed-supervision training, the model is trained for 200 epochs, consisting of 100 epochs of supervised warm-up and 100 epochs of mixed training. All experiments follow consistent training strategies to ensure fair performance evaluation.

\subsection{Comparison Experiments}

To comprehensively and objectively evaluate UniCD's performance, this study selected twenty-seven representative methods across three paradigms, with nine methods for each category: (i) Supervised methods: SNUNet \cite{24snunet}, ChangeFormer \cite{25transformer}, DMINet \cite{26dminet}, SAM-CD \cite{13samcd}, CDMask \cite{27cdmask}, ChangeMamba \cite{12changemamba}, MetaCD  \cite{28metacd}, MDENet \cite{29mdenet}, and DA²Net \cite{30da2net}. (ii) Weakly-supervised methods: WCDNet \cite{14wcdnet}, SEAM \cite{15seam}, MS-Former \cite{msformer}, MCTFormer \cite{mctformer}, BGMix \cite{bgmix}, FCD-GAN \cite{9fcdgan}, TransWCD \cite{zhao2025transwcd}, CS-WSCD \cite{16cswscd}, and S2C \cite{s2c}. (iii) Unsupervised methods: DCVA \cite{dcva}, DSFA \cite{dsfa}, KPCA \cite{kpca}, CDRL \cite{cdrl}, SiROC \cite{siroc}, I3PE \cite{i3pe}, FCD-GAN \cite{9fcdgan}, AnyChange \cite{anychange} and S2C \cite{dings2c}.

\textbf{Mixed-Supervision CD Results.} To evaluate UniCD under different heterogeneous-supervision configurations, we conduct experiments on LEVIR-CD, WHU-CD, and CLCD using five supervision ratios ($\mathcal D_s:\mathcal D_w:\mathcal D_u=6:2:2$, $4:3:3$, $1:1:1$, $1:1:0$, and $1:0:1$). As shown in Table~\ref{mixed_ratio}, directly mixing different supervision types does not improve performance over supervised-only training, indicating that complementary supervision signals cannot be fully exploited through naive joint training. Incorporating either SCPP or MCR improves performance to varying degrees, while their combination achieves the best results among mixed-training configurations across all three datasets and supervision ratios. These results demonstrate the effectiveness of UniCD in exploiting complementary information from heterogeneous supervision signals, providing a promising solution for better utilizing diverse annotated and unlabeled data in change detection.

\textbf{Individual Supervision CD Results.} Tables \ref{supervised_comparison}, \ref{weak_comparison}, and \ref{unsupervised_comparison} present the performance comparisons of UniCD under fully supervised, weakly supervised, and unsupervised settings on LEVIR-CD, WHU-CD, and CLCD. UniCD consistently achieves the best F1 and IoU scores across all three datasets and supervision paradigms. Qualitative comparisons on LEVIR-CD in Fig.~\ref{i4} further demonstrate more accurate change localization with fewer false positives and false negatives, validating the effectiveness of UniCD under diverse supervision conditions.

\begin{table}[!t]
\scriptsize
\centering
\renewcommand{\arraystretch}{1.1}
\setlength{\tabcolsep}{2pt}
\caption{\textbf{Weakly supervised results} (\%) on LEVIR-CD, WHU-CD, and CLCD. The best values are highlighted in bold.}
\label{weak_comparison}
\begin{tabular}{l|ccc|ccc|ccc}
\toprule[1.5pt]
\multirow{2}{*}{\textbf{Model}}
& \multicolumn{3}{c|}{\textbf{LEVIR-CD}}
& \multicolumn{3}{c|}{\textbf{WHU-CD}}
& \multicolumn{3}{c}{\textbf{CLCD}} \\
\cline{2-10}
& \textbf{F1} & \textbf{IoU} & \textbf{OA}
& \textbf{F1} & \textbf{IoU} & \textbf{OA}
& \textbf{F1} & \textbf{IoU} & \textbf{OA} \\
\midrule[1pt]
WCDNet$_{\scriptscriptstyle\mathrm{ACCV}}$
& 34.45 & 20.81 & 85.46 & 48.90 & 32.36 & 90.56 & 25.66 & 14.72 & 92.11 \\

SEAM$_{\scriptscriptstyle\mathrm{CVPR}}$
& 41.38 & 26.09 & 88.99 & 50.10 & 33.42 & 94.89 & 51.01 & 34.23 & 92.29 \\

MS-Former$_{\scriptscriptstyle\mathrm{CAIP}}$
& 64.64 & 47.76 & 96.36 & 65.68 & 48.89 & 95.53 & 37.17 & 22.83 & 87.18 \\

MCTFormer$_{\scriptscriptstyle\mathrm{CVPR}}$
& 64.55 & 44.66 & 95.45 & 55.99 & 38.87 & 93.01 & 34.39 & 20.77 & 90.10 \\

BGMix$_{\scriptscriptstyle\mathrm{AAAI}}$
& 52.03 & 35.16 & 96.84 & 59.43 & 42.28 & 96.29 & 33.90 & 20.41 & 91.17 \\

FCD-GAN$_{\scriptscriptstyle\mathrm{TPAMI}}$
& 55.42 & 38.33 & 93.79 & 50.74 & 33.98 & 96.13 & 25.88 & 14.86 & 92.92 \\

TransWCD$_{\scriptscriptstyle\mathrm{TGRS}}$
& 63.44 & 46.46 & 95.97 & 70.01 & 53.87 & 96.44 & 47.70 & 31.32 & 92.78 \\

CS-WSCD$_{\scriptscriptstyle\mathrm{TGRS}}$
& 64.14 & 47.21 & 96.87 & 73.05 & 57.53 & 97.08 & 49.96 & 33.30 & 91.10 \\

S2C$_{\scriptscriptstyle\mathrm{CVPR}}$
& 65.08 & 48.22 & 96.35 & 73.40 & 58.01 & \textbf{97.47} & 50.84 & 34.08 & 91.54 \\

\midrule[1pt]
\textbf{UniCD}
& \textbf{73.84} & \textbf{58.54} & \textbf{97.29}
& \textbf{74.05} & \textbf{58.79} & 97.38
& \textbf{58.05} & \textbf{40.90} & \textbf{93.66} \\
\bottomrule[1.5pt]
\end{tabular}
\end{table}

\begin{table}[!t]
\scriptsize
\centering
\renewcommand{\arraystretch}{1.1}
\setlength{\tabcolsep}{2pt}
\caption{\textbf{Unsupervised results} (\%) on LEVIR-CD, WHU-CD, and CLCD. The best values are highlighted in bold.}
\label{unsupervised_comparison}
\begin{tabular}{l|ccc|ccc|ccc}
\toprule[1.5pt]
\multirow{2}{*}{\textbf{Model}}
& \multicolumn{3}{c|}{\textbf{LEVIR-CD}}
& \multicolumn{3}{c|}{\textbf{WHU-CD}}
& \multicolumn{3}{c}{\textbf{CLCD}} \\
\cline{2-10}
& \textbf{F1} & \textbf{IoU} & \textbf{OA}
& \textbf{F1} & \textbf{IoU} & \textbf{OA}
& \textbf{F1} & \textbf{IoU} & \textbf{OA} \\
\midrule[1pt]
DCVA$_{\scriptscriptstyle\mathrm{TGRS}}$
& 12.21 & 6.50 & 56.57 & 9.06 & 4.75 & 54.46 & 17.35 & 9.50 & 55.45 \\

DSFA$_{\scriptscriptstyle\mathrm{TGRS}}$
& 10.09 & 5.32 & 72.64 & 10.00 & 5.26 & 74.04 & 18.32 & 10.08 & 57.81 \\

KPCA-MNet$_{\scriptscriptstyle\mathrm{TCYB}}$
& 9.05 & 4.74 & 62.04 & 11.91 & 6.33 & 72.26 & 17.16 & 9.38 & 70.06 \\

CDRL$_{\scriptscriptstyle\mathrm{RSL}}$
& 10.30 & 5.43 & 64.88 & 8.49 & 4.43 & 74.54 & 12.26 & 6.54 & 49.49 \\

SiROC$_{\scriptscriptstyle\mathrm{TGRS}}$
& 9.58 & 5.03 & 78.91 & 9.37 & 4.92 & 81.90 & 19.79 & 11.00 & 80.87 \\

I3PE$_{\scriptscriptstyle\mathrm{ISPRS\ JPRS}}$
& 30.87 & 18.26 & 86.62 & 23.52 & 13.33 & 79.97 & 34.44 & 20.78 & 85.87 \\

FCD-GAN$_{\scriptscriptstyle\mathrm{TPAMI}}$
& 11.08 & 5.86 & 71.76 & 9.71 & 5.10 & 77.53 & 34.86 & 21.12 & 83.63 \\

AnyChange$_{\scriptscriptstyle\mathrm{NeurIPS}}$
& 24.71 & 14.09 & 76.18 & 22.86 & 12.90 & 76.15 & 27.39 & 15.87 & 80.88 \\

S2C$_{\scriptscriptstyle\mathrm{AAAI}}$
& 34.07 & 20.53 & 88.46 & 36.02 & 21.96 & 90.58 & 47.03 & 30.75 & 92.26 \\

\midrule[1pt]
\textbf{UniCD}
& \textbf{46.06} & \textbf{29.92} & \textbf{92.19}
& \textbf{39.90} & \textbf{24.92} & \textbf{90.42}
& \textbf{51.19} & \textbf{34.40} & \textbf{92.57} \\
\bottomrule[1.5pt]
\end{tabular}
\end{table}

\begin{figure}[!t]
\centering
\includegraphics[scale = 0.37]{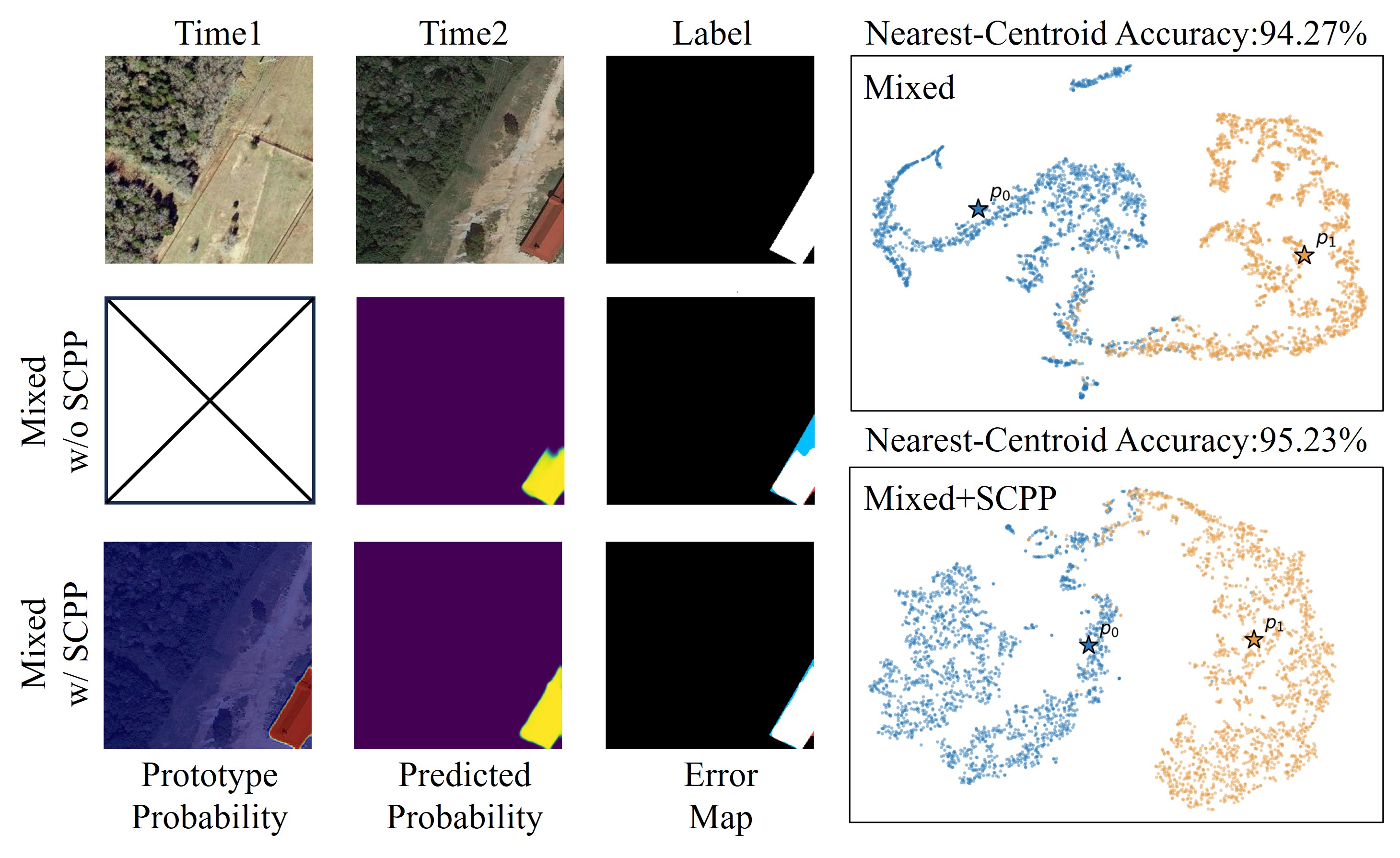}
\caption{\textbf{Visualization of SCPP} effects and t-SNE feature distributions on LEVIR-CD. Compared with mixed training without SCPP, incorporating SCPP improves change prediction and feature discrimination, increasing nearest-centroid accuracy from 94.27\% to 95.23\%.}
\label{visscpp}
\end{figure}

\subsection{Discussion}

\textbf{Analysis of SCPP.} As shown in Fig.~\ref{visscpp}, incorporating SCPP into mixed training improves prototype responses and change predictions, leading to fewer prediction errors. The t-SNE visualizations further demonstrate clearer separation between changed and unchanged features, with nearest-centroid accuracy increasing from 94.27\% to 95.23\%, confirming the effectiveness of SCPP in enhancing feature discrimination.

\begin{figure}[!t]
\centering
\includegraphics[scale = 0.36]{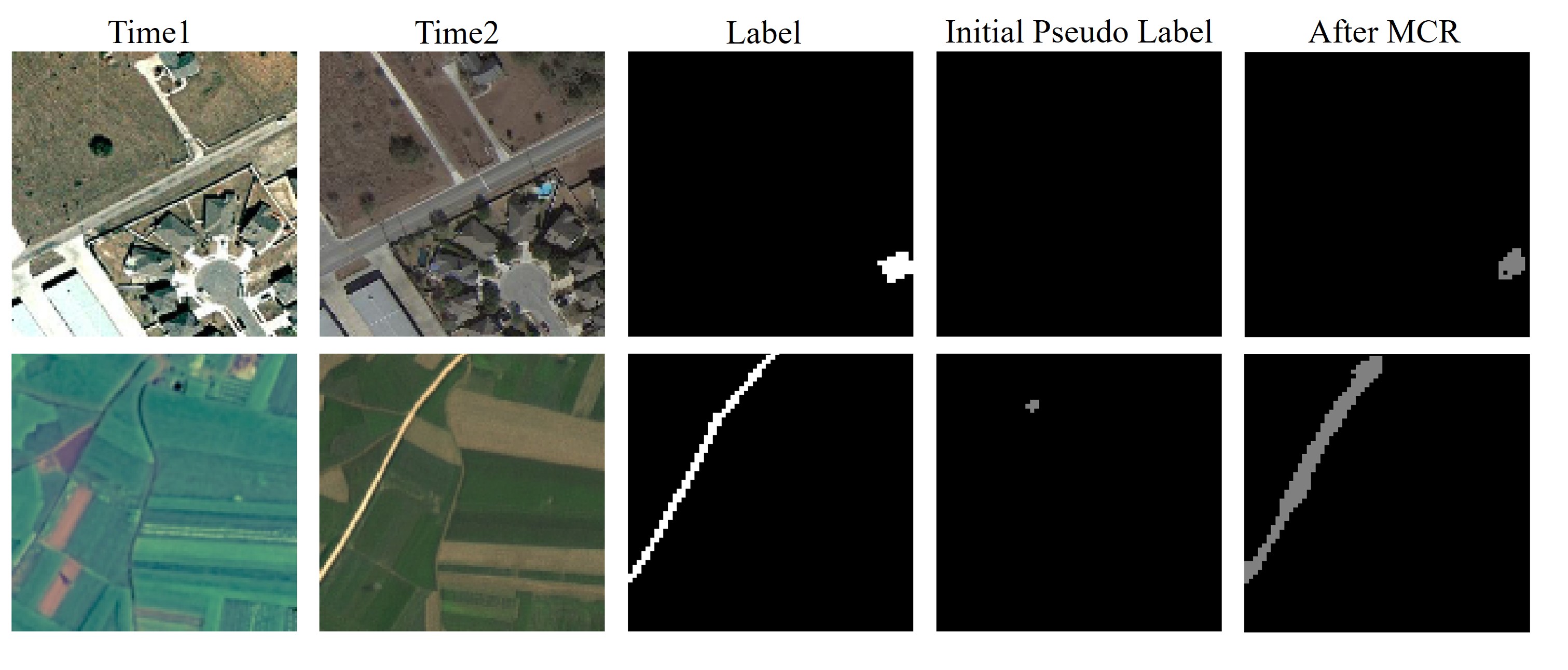}
\caption{\textbf{Visualization of MCR} effects on pseudo-label refinement. Compared with the initial pseudo labels, MCR produces higher-quality pseudo labels for model optimization.}
\label{vismcr}
\end{figure}

\textbf{Analysis of MCR.} As shown in Fig.~\ref{vismcr}, MCR improves pseudo-label quality by producing more accurate and spatially complete change predictions. Compared with the initial pseudo labels, the refined results exhibit better correspondence with the ground-truth change regions, particularly for the line-shaped changes in the second example. These results demonstrate the effectiveness of MCR in improving pseudo-label reliability and providing more informative supervision for model optimization.

\begin{table}[!t]
\footnotesize
\centering
\renewcommand{\arraystretch}{0.6}
\setlength{\tabcolsep}{1.5pt}
\caption{\textbf{Complexity comparison} with representative change detection methods under different supervision paradigms.}
\label{complexity}
\begin{tabular}{l|l|cc}
\toprule
\textbf{Learning Setting} & \textbf{Method}
& \textbf{Params.(M)}
& \textbf{Inference Time (ms)} \\
\midrule
\multirow{4}{*}{Supervised}
& ChangeFormer  & 41.01 & 17.323 \\
& CDMask        & 33.52 & 47.208 \\
& ChangeMamba   & 49.94 & 37.822 \\
& MDENet        & 60.77 & 15.846 \\
\midrule
\multirow{4}{*}{Weakly Sup.}
& MCTFormer     & 21.79 & 4.995 \\
& CS-WSCD       & 99.93 & 3.465 \\
& S2C           & 106.12 & 5.107 \\
& BGMix         & 30.22 & 5.844 \\
\midrule
\multirow{2}{*}{Unsupervised}
& CDRL          & 54.54 & 8.858 \\
& I3PE          & 24.93 & 7.288 \\
\midrule
\multirow{1}{*}{Multi-paradigm}
& FCD-GAN       & 40.82 & 6.248 \\
\midrule
\rowcolor{gray!20}
\multirow{1}{*}{Unified}
& UniCD
& 32.75
& 10.764 \\
\bottomrule
\end{tabular}
\end{table}


\textbf{Complexity Analysis.} Table~\ref{complexity} compares the model complexity of representative methods under different supervision paradigms. UniCD contains 32.75M parameters, fewer than FCD-GAN and comparable to individual-supervision methods, demonstrating its ability to support heterogeneous-supervision learning with moderate model complexity.

\section{Conclusion}
We propose UniCD, a unified framework for remote sensing change detection that supports supervised, weakly supervised, unsupervised, and mixed-supervision learning within a shared network. To facilitate heterogeneous supervision, we develop Supervision-Anchored Cross-Granularity Prototype Propagation (SCPP) to transfer class-specific knowledge from pixel-level annotations, and Multi-source Consistency Rectification (MCR) to improve pseudo-label quality through reliability screening and selective recovery. Experiments on LEVIR-CD, WHU-CD, and CLCD validate the effectiveness of UniCD across different supervision settings. UniCD enables pixel-level annotations, image-level labels, and unlabeled samples to be jointly exploited within a unified learning framework, effectively bridging heterogeneous supervision paradigms and facilitating the complementary use of diverse supervision signals. These results demonstrate the potential of UniCD as a unified solution to heterogeneous supervision in RSCD.
{
    \small
    \bibliographystyle{ieeenat_fullname}
    \bibliography{main}
}


\end{document}